\pdfoutput=1

\documentclass[11pt]{article}
\usepackage[dvipsnames,table,xcdraw]{xcolor}

\usepackage[]{EMNLP2022}

\usepackage{times}
\usepackage{latexsym}
\usepackage{booktabs}
\usepackage{kotex}
\usepackage{CJKutf8}
\usepackage[utf16]{}

\newcommand{\tabitem}{~~\llap{\textbullet}~~}

\usepackage[T1]{fontenc}

\usepackage[utf8]{inputenc}

\usepackage{microtype}

\usepackage[nointegrals]{wasysym}
\usepackage{graphicx}
\usepackage{multirow}
\usepackage{array}
\usepackage{lipsum}
\usepackage{amsmath}
\usepackage{subcaption}
\usepackage{diagbox}
\usepackage{slashbox}

\usepackage{tikz}

\newenvironment{myfont}{\fontfamily{qcr}\selectfont}{\par}
\DeclareTextFontCommand{\code}{\myfont}

\newcolumntype{?}{!{\vrule width 1.2pt}}

\usepackage{makecell}
\usepackage{kotex}
\usepackage{CJKutf8}

\usepackage{inconsolata}

\title{Translating Hanja Historical Documents to\\Contemporary Korean and English}

\author
{
    Juhee Son$^{1*}$, Jiho Jin$^{1*}$, Haneul Yoo$^1$ , JinYeong Bak$^2$, Kyunghyun Cho$^{3,4}$, Alice Oh$^1$ \\
    $^1$KAIST, $^2$Sungkyunkwan University, $^3$New York University, $^4$Genentech \\
    \texttt{\{sjh5665, jinjh0123, haneul.yoo\}@kaist.ac.kr}, \\
    \texttt{jy.bak@skku.edu}, \texttt{kyunghyun.cho@nyu.edu}, \texttt{alice.oh@kaist.edu}
}

\begin{document}
\maketitle
\def\thefootnote{*}\footnotetext{Equal contribution.}
\renewcommand{\thefootnote}{\arabic{footnote}}
\begin{abstract}

The Annals of Joseon Dynasty (AJD) contain the daily records of the Kings of Joseon, the 500-year kingdom preceding the modern nation of Korea.
The Annals were originally written in an archaic Korean writing system, `Hanja', and were translated into Korean from 1968 to 1993.
The resulting translation was however too literal and contained many archaic Korean words; thus, a new expert translation effort began in 2012. Since then, the records of only one king have been completed in a decade.
In parallel, expert translators are working on English translation, also at a slow pace and produced only one king's records in English so far.
Thus, we propose H2KE, a neural machine translation model, that translates historical documents in Hanja to more easily understandable Korean and to English.
Built on top of multilingual neural machine translation, H2KE learns to translate a historical document written in Hanja, from both a full dataset of outdated Korean translation and a small dataset of more recently translated contemporary Korean and English.
We compare our method against two baselines:
a recent model that simultaneously learns to restore and translate Hanja historical document
and a Transformer based model trained only on newly translated corpora.
The experiments reveal that our method significantly outperforms the baselines in terms of BLEU scores for both contemporary Korean and English translations.
We further conduct extensive human evaluation which shows that our translation is preferred over the original expert translations by both experts and non-expert Korean speakers.

\end{abstract}

\section{Introduction}

\begin{table}[t]
\small
\renewcommand*{\arraystretch}{1.1}
\begin{tabular}{c|m{5cm}}
\toprule
\textbf{Hanja}  &   改淸州牧爲西原縣. 以\colorbox[HTML]{D0D4D8}{劇賊}胎生邑, \colorbox[HTML]{ACBADB}{降號}也.  \\ \midrule
\multirow{2}{*}{\begin{tabular}[c]{@{}c@{}}\textbf{Original} \\ \textbf{Korean}\\ \textbf{Translation} \\ \textbf{(oKo)}\end{tabular}}   & 청주목을 서원현으로 고쳤다. \colorbox[HTML]{D0D4D8}{극적}이 태생한 고을은 \colorbox[HTML]{ACBADB}{강호}하기 때문이다.
 \\  & Eng.) Cheongju-mok was renamed Seowon-hyeon. It is because the town gets \colorbox[HTML]{ACBADB}{gangho} if \colorbox[HTML]{D0D4D8}{geukjeok} is born. \\ \midrule
\multirow{2}{*}{\begin{tabular}[c]{@{}c@{}}\textbf{Contemporary}\\ \textbf{Korean} \\ \textbf{Translation} \\ \textbf{(cKo)} \end{tabular}} & 청주목을 서원현으로 고쳤다. \colorbox[HTML]{D0D4D8}{극악한} \colorbox[HTML]{D0D4D8}{역적}이 태어난 고을이므로 \colorbox[HTML]{ACBADB}{읍호를} \colorbox[HTML]{ACBADB}{강등}한 것이다.  \\  & Eng.) Cheongju-mok was renamed Seowon-hyeon. Since it is a town where a \colorbox[HTML]{D0D4D8}{vicious traitor} was born, the town was \colorbox[HTML]{ACBADB}{demoted}.\\ \bottomrule
\end{tabular}
\caption{An example from the Annals of Joseon Dynasty. We show the original Hanja sentence and the original Korean human translation which contains archaic words indicated in color box. The contemporary Korean translation replaces the archaic words with words and phrases understood by present-day Korean speakers.}
\label{tab:ex_data}
\end{table}

Historical documents written in an archaic language should be translated into a modern language.
Most of the Korean historical documents are written in Hanja, the main written language in Korea before the 20-th century. Hanja is an archaic language based on the old Chinese writing system, and although there is a large overlap in characters, it is different from both Chinese and Korean.
\emph{The Annals of Joseon Dynasty} (AJD), the representative historical records of Joseon (1392 - 1910), originally written in Hanja, was translated into Korean from 1968 to 1993 by expert translators commissioned by the Korean government.
Non-expert Korean speakers however have trouble understanding these original translations of the AJD because they contain many archaic Hanja-based words, often hard-to-understand transliterations.
The Institute for the Translation of Korean Classics (ITKC) recognizes this problem and is re-translating the entire AJD with modern-style writing
(Table  \ref{tab:ex_data}). 
This re-translation process is expected to take 22 years with 12 to 15 expert translators. 
Simultaneously, the National Institute of Korean History (NIKH) has been translating AJD into English since 2012, which is also expected to take about two more decades.

Machine translation can accelerate the translation process. The challenge is the limited availability of parallel corpora between Hanja to contemporary Korean as well as English. 
Only one annal of the 24 kings of the Joseon Dynasty was newly translated into Korean and English. This is not a sufficient amount to train a full machine translation model. 
To address this low-resource problem, we adopt a multilingual translation approach that jointly learns to translate between Hanja, outdated original Korean, contemporary Korean  and English, expecting positive transfer of knowledge among these languages.

We present a multilingual neural machine translation model that translates Hanja historical documents to contemporary Korean, to which we refer as \textbf{H2KE}.
By exploiting extra resources, H2KE performs significantly better translation of Hanja into contemporary Korean than other approaches that rely solely on the parallel corpus from the newly translated Korean and Hanja. 
We measure the perplexity with a large-scale language model trained on contemporary Korean, called KoGPT \cite{kakaobrain2021kogpt}, to show that translations from our model  are more similar to contemporary Korean than the old Korean translations from the original translation effort. 
These results are further confirmed by human evaluation, where both experts and non-experts prefer our model's translation over the original translation in old Korean.
Using H2KE, we translated the remaining AJD to contemporary Korean as well as English and are releasing it publicly at \url{https://juheeuu.github.io/h2ke-demo}.

Our main contributions include:
\begin{itemize}
\item{We propose a transfer learning method for translating AJD to contemporary Korean and English with a small training corpus.}

\item{We conduct thorough human evaluation, where experts find that our generated translations are more accurate and fluent than the original expert translations, and non-expert Korean speakers choose our translations as more easily understandable compared to the original translations.}

\item{We translate the entire AJD to modern Korean and English and publicly release the translations for easier access to the resources.}

\end{itemize}

\section{Background}

\subsection{Neural Machine Translation for the Annals of the Joseon Dynasty}

To translate AJD with the neural network, \citet{9125904} propose a new subword tokenization method called share-vocabulary-and-entity-restriction byte-pair encoding.
\citet{kang2021restoring} present a multi-task learning approach that simultaneously restores and translates historical documents. 
For the restoration task, they use the untranslated Diaries of the Royal Secretariat (DRS) which is another Korean historical corpus written in Hanja.
For translation, they only focus on translating Hanja into old Korean using the outdated AJD corpus.
In contrast to these earlier approaches, our approach supports both translation into contemporary Korean and into English, while benefiting from the larger Hanja-old Korean parallel corpus.

\subsection{The Annals of the Joseon Dynasty}

\emph{The Annals of the Joseon Dynasty} (AJD), also called \emph{the Veritable Records of the Joseon Dynasty}, is an old and vast volume of historical documents from Joseon Dynasty which ruled the Korean peninsula from 1392 to 1864.
It records 472 years of the 25 rulers' reigns of the Joseon Dynasty.
It covers diverse historical events and is known to exhibit high integrity and credibility in its description of these events, making it invaluable as a historical record.\thinspace\footnote{
The description for AJD is based on Korean Cultural Heritage Administration (\url{https://www.cha.go.kr/}).
}
The dataset is available at `the Veritable Records of the Joseon Dynasty'\thinspace\footnote{
 \url{http://sillok.history.go.kr/}
} 
run by the National Institute of Korean History (NIKH).
AJD was originally written in Hanja, the writing system of ancient Korea, consisting of totally different characters and syntactic structures from contemporary Korean.
Hanja had stemmed from traditional Chinese, but the lexical, semantic, and syntactic characteristics had changed to reflect the cultural differences between the Joseon Dynasty and other ancient Kingdoms of China.

\begin{table*}[t]
\centering
\newcolumntype{M}[1]{>{\centering\arraybackslash}m{#1}}
\begin{tabular}{l|c|M{1.25cm}M{1.25cm}M{1.25cm}M{1.25cm}|M{1.3cm}|r}
\toprule
 \textbf{Annals of} & \textbf{Reign} & \textbf{Hanja} & \textbf{oKo} & \textbf{cKo} & \textbf{English} & \textbf{\# of sentences}& \multicolumn{1}{c}{\textbf{Ratio (\%)}}\\
\midrule 
\textbf{Joseon Dynasty}& 1392-1910 & \Circle & \Circle &         & & \multicolumn{1}{r|}{359,726} & 100.0 \\
\textbf{22\textsuperscript{th} King Jeongjo}&1776-1799& \Circle & \Circle & \Circle & & \multicolumn{1}{r|}{14,356}  & 3.9 \\
\textbf{4\textsuperscript{th} King Sejong}&1418-1449& \Circle & \Circle &  & \Circle & \multicolumn{1}{r|}{26,227} & 7.2 \\
\bottomrule
\end{tabular}
\caption{
Statistics of our dataset. 
For the entire AJD, there are \textlangle{}Hanja, oKo\textrangle{} pairs.
For the Annals of King Jeongjo, we also have contemporary Korean translations, and for the Annals of King Sejong, we have the English translations. 
The last column indicates the ratio of each dataset on the basis of the total AJD.
}
\label{tab:dataset}
\end{table*}

\subsection{Translated Datasets}

AJD was initially translated from Hanja to Korean during 1968 - 1993, and the dataset was uploaded and publicly released by the Institute for the Translation of Korean Classics (ITKC).\footnote{
Both the original translation of AJD and the new translation of AKJ are available at \url{https://db.itkc.or.kr/}.
}
These original translations include numerous outdated Hanja-based words, often transliterations. These words are often not easily understood by  contemporary Korean speakers, or are simply incorrect in the context they appear.
To correct those and other errors and also to improve the overall readability, ITKC  launched a project for modernizing the translation of AJD in 2011.
\emph{The Annals of the 22-nd King Jeongjo} (AKJ) was the first one to be translated between 2012 and 2016.
Throughout this paper, we  refer to the original translation as \emph{oKo} and the new contemporary translation as \emph{cKo}.
For the globalization of AJD, listed as UNESCO's Memory of the World, and Korean history, NIKH has been translating AJD into English, in parallel to the effort by ITKC, since 2013.
\emph{The Annals of the 4-th King Sejong} (AKS) has been translated so far, and it is available from \url{http://esillok.history.go.kr/}.
These translation projects are expected to take two decades.

In Table \ref{tab:dataset} we list these corpora and their statistics.
As discussed earlier, the corpora for contemporary Korean and English are substantially smaller than those for old Korean.

\section{Method}

\textbf{H2KE} is a model that learns to translate historical documents written in \textbf{H}anja to contemporary \textbf{K}orean and \textbf{E}nglish. We use the multilingual neural machine translation (MNMT) approach, which enables translation between multiple languages with a single model \cite{johnson-etal-2017-googles,firat2016multi}.

\paragraph{Multilingual Translation Approach.}

\begin{figure}[t]
    \centering
    \includegraphics[width=0.95\linewidth]{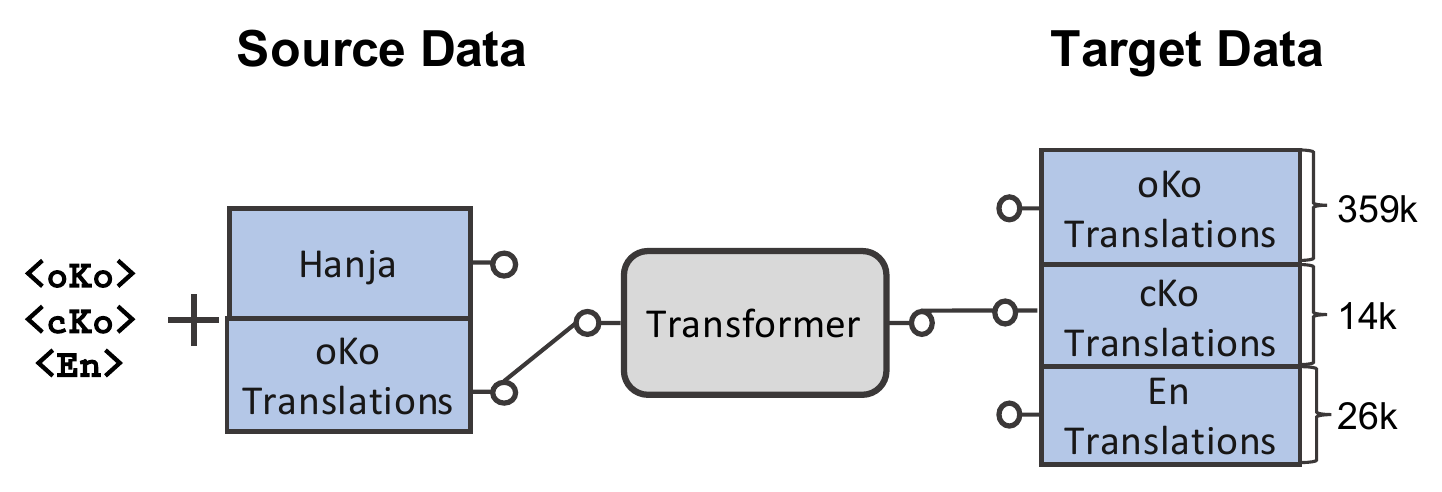}
    \caption{H2KE works with multiple language pairs by
    appending a source sentence with a target language token during training and inference.}
    \label{fig:model}
\end{figure}

Our dataset consists of \textlangle{}source, target\textrangle{} pairs of \textlangle{}Hanja, oKo\textrangle{}, \textlangle{}Hanja, cKo\textrangle{}, \textlangle{}Hanja, English\textrangle{}, \textlangle{}oKo, cKo\textrangle{}, and \textlangle{}oKo, English\textrangle{}.
We append a special target-language token (either \code{<oKo>}, \code{<cKo>}, or \code{<En>}) in front of each source sentence.
We train a model using 
all these examples shuffled randomly by presenting one pair of sentences at a time.
Figure \ref{fig:model} illustrates the overall translation pipeline.
With this approach, the model can benefit from the large amount of \textlangle{}Hanja, oKo\textrangle{} to improve the translation quality of the lower-resource target language pairs, \textlangle{}Hanja, cKo\textrangle{} and \textlangle{}Hanja, English\textrangle{}.

\paragraph{Training and Inference.}

We use the Transformer model \cite{vaswani2017attention} to implement H2KE. We optimize the following loss for training:

\begin{equation}
\begin{split}
\mathcal{L}= & -
\frac{1}{N} 
\sum_{n=1}^N 
\sum_{t=1}^{T_n}{\log{p_{\theta}(y^{(n)}_t|y^{(n)}_{<t},x^{(n)},tok^{(n)})}}.
\end{split}
\label{eq:loss}
\end{equation}
 There are $N$ training examples, and each example is tagged with the target side language using  $tok^{(n)} \in$ \{\code{<oKo>}, \code{<cKo>}, \code{<En>}\}.

For generation, we use beam search 
and
translate the Hanja sentences to the language specified by the target language token. We generate and evaluate sentences in target languages, English (EN) and contemporary Korean (cKo), with either Hanja or original Korean translation (oKo) as source sentences.

\section{Experiments and Results}
\begin{table*}[t]
\centering
\begin{tabular}{c|c|lll|ccccc@{}}
\toprule

&\multirow{3}{*}{\textbf{Model}}& \multicolumn{1}{c}{\textbf{All}} & \multicolumn{1}{c}{\textbf{Jeongjo}} & \multicolumn{1}{c|}{\textbf{Sejong}} & \multicolumn{5}{c}{\textbf{BLEU}} \\&& HJ & HJ/oKo & HJ/oKo& \multicolumn{1}{l|}{HJ} & \multicolumn{1}{l}{HJ} & \multicolumn{1}{l|}{oKo} & \multicolumn{1}{l}{HJ} & \multicolumn{1}{l}{oKo} \\
 && \multicolumn{1}{r}{→oKo}& →cKo & →EN& \multicolumn{1}{r|}{→oKo} & \multicolumn{1}{r}{→cKo}& \multicolumn{1}{r|}{→cKo}&\multicolumn{1}{r}{→EN} & \multicolumn{1}{r}{→EN} \\ \midrule

\multirow{1}{*}{(A)}&\multicolumn{1}{l|}{Papago}& &\multicolumn{1}{c}{} & & \multicolumn{1}{c|}{-}&\multicolumn{1}{c}{11.10}& \multicolumn{1}{c|}{-} &3.59&4.49\\  \midrule
\multirow{3}{*}{(B)}&\multicolumn{1}{p{1.7cm}|}{\citeauthor{kang2021restoring}} & \multicolumn{1}{c}{\Circle}  & &  & \multicolumn{1}{c|}{41.56}&-& \multicolumn{1}{c|}{-}&-&-\\ 
&\multicolumn{1}{l|}{H2KE-base}&\multicolumn{1}{c}{\Circle} & & & \multicolumn{1}{c|}{46.23}&-& \multicolumn{1}{c|}{-} &-&-\\ 
&\multicolumn{1}{l|}{H2KE-big}&\multicolumn{1}{c}{\Circle} & & & \multicolumn{1}{c|}{\textbf{47.57}}&-& \multicolumn{1}{c|}{-} &-&-\\ \midrule

\multirow{2}{*}{(C)}
&\multicolumn{1}{l|}{H2KE-big}& &\multicolumn{1}{c}{\Circle} & & \multicolumn{1}{c|}{-}&17.63& \multicolumn{1}{c|}{21.43} &-&-\\ 
&\multicolumn{1}{l|}{H2KE-big}&\multicolumn{1}{c}{\Circle}&\multicolumn{1}{c}{\Circle} & & \multicolumn{1}{c|}{46.76}&\textbf{46.44}& \multicolumn{1}{c|}{\textbf{45.76}} &-&-\\  \midrule
\multirow{2}{*}{(D)} &\multicolumn{1}{l|}{H2KE-big}&&&\multicolumn{1}{c|}{\Circle}  & \multicolumn{1}{c|}{-}&-& \multicolumn{1}{c|}{-} &11.92&12.36\\ 
&\multicolumn{1}{l|}{H2KE-big}&\multicolumn{1}{c}{\Circle}&&\multicolumn{1}{c|}{\Circle}  & \multicolumn{1}{c|}{46.23}&-& \multicolumn{1}{c|}{-} &\textbf{25.23}&24.50\\ \midrule
(E) &\multicolumn{1}{l|}{H2KE-big}&\multicolumn{1}{c}{\Circle}&\multicolumn{1}{c}{\Circle}&\multicolumn{1}{c|}{\Circle}  & \multicolumn{1}{c|}{46.58}&46.11& \multicolumn{1}{c|}{\textbf{45.76}} &24.62&\textbf{24.59}\\ 
\bottomrule
\end{tabular}
\caption{
Test results of our model on different training dataset combinations.
The circle indicates the king of annals and the language pair of the data for training.
The BLEU score of one target language can be measured on the different source languages.
}
\label{tab:bleu}
\end{table*}

\subsection{Data Preprocessing and Training Settings}

We use the unigram language model tokenizer \cite{kudo2018subword} provided by Google's SentencePiece library.\footnote{
\url{https://github.com/google/sentencepiece}
}
In order to use one shared vocabulary between source and target languages, we tokenize the entire corpus together, including Hanja, oKo, cKo and EN. 
We limit the size of the vocabulary to 32K. 
The out-of-vocabulary tokens are replaced with UNK (unknown) tokens. 
We use the hyperparameters recommended by \citet{vaswani2017attention}.
We train and evaluate models using Fairseq \cite{ott2019fairseq}.
We average the five best checkpoints on validation data to obtain the final model to be tested on the test set.

\subsection{Translation Quality}

We train models with different dataset combinations and measure the BLEU score \cite{papineni2002bleu}.
To measure the Korean BLEU score, we follow the protocol from WAT 2019 \cite{nakazawa2019proceedings} and use Mecab-ko\thinspace\footnote{
\url{https://bitbucket.org/eunjeon/mecab-ko/}
} tokenizer and Sacrebleu \cite{post2018call}.
For English, we use Sacrebleu.

Table \ref{tab:bleu} shows the BLEU score for each case.
Overall, utilizing \textlangle{}Hanja, oKo\textrangle{} pairs brings significant improvement in low-resource translations (to cKo or EN).
However, there exist performance degradations when adding the unrelated target language pairs to the translation from Hanja.
Since the encoder already learns expressive representations for Hanja from the plenty of training samples, inserting pairs with different target languages rather hinders the representation learning of the source language, Hanja.

\paragraph{A Commercial Translation Engine.}

We first compare our models to the Korean-specialized commercial translation service, called Papago \cite{lee2016papago}.
Although Papago was never trained to translate Hanja into modern Korean nor into English, we can force it to do so by asking it to translate from Taiwanese Mandarin (zh-TW) which shares a large set of characters with Hanja.
According to the row (A) in Table \ref{tab:bleu}, the commercial translation system, Papago, simply fails to properly translate Hanja documents, evident from significantly low BLEU in both contemporary Korean and English.

\paragraph{Original Korean Translation.} 

Although there is no preceding work on translating Hanja into either contemporary Korean or English, \citet{kang2021restoring} had recently demonstrated the effectiveness of neural machine translation for translating Hanja into old Korean. We thus compare our approach against theirs in Hanja-Old Korean translation. For fair comparison, 
we only use the \textlangle{}Hanja, oKo\textrangle{} corpus and train a H2KE-base with only 65M parameters. 

As shown in the row group (B) in Table \ref{tab:bleu},
the proposed H2KE-base achives 5 BLEU scores higher than \citet{kang2021restoring}.
We attribute this improvement to the vocabulary sharing strategy and the use of the transformer.
Without vocabulary sharing, the model showed 45.09 BLEU score. 
When we try a larger model, H2KE-big with 213M parameters, we achieve even better translation quality. We thus stick to H2KE-big in the rest of the experiments.

\paragraph{Contemporary Korean Translation.} 

The first row in the row group (C) of Table \ref{tab:bleu} shows that the model trained with only a small amount of \textlangle{}Hanja, cKo\textrangle{} and \textlangle{}oKo, cKo\textrangle{} pairs result in low BLEU scores.
However, adding the \textlangle{}Hanja, oKo\textrangle{} parallel corpus dramatically improves translation quality for the cKo translations, evident from 20-30 BLEU scores increase. 
This confirms the effectiveness of multilingual training which we hypothesized earlier.

When we take the original Korean (oKo) as translation and compare it against the ground truth contemporary Korean (cKo) as reference, we obtain the BLEU score of 39.74. This score is 
lower than that of the H2KE's cKo translation. This strongly suggests that the generated translations from our system are more similar to the cKo than the expert's ground truth oKo translations, fulfilling the goal of producing a machine translation system for contemporary Korean.

\paragraph{English Translation.}

According to the result in the row (D) in Table \ref{tab:bleu}, 
we observe a similar trend when we use H2KE for translating Hanja into English.
We gain significant improvement in translation quality by including  the \textlangle{}Hanja, oKo\textrangle{} corpus during training.
Finally in the final row (E) of Table \ref{tab:bleu}, we demonstrate that a single H2KE-big model can be trained on all the corpora and can translate Hanja into both old Korean, contemporary Korean and English competitively. 


\begin{figure}[t]
    \centering
    \includegraphics[width=0.95\linewidth]{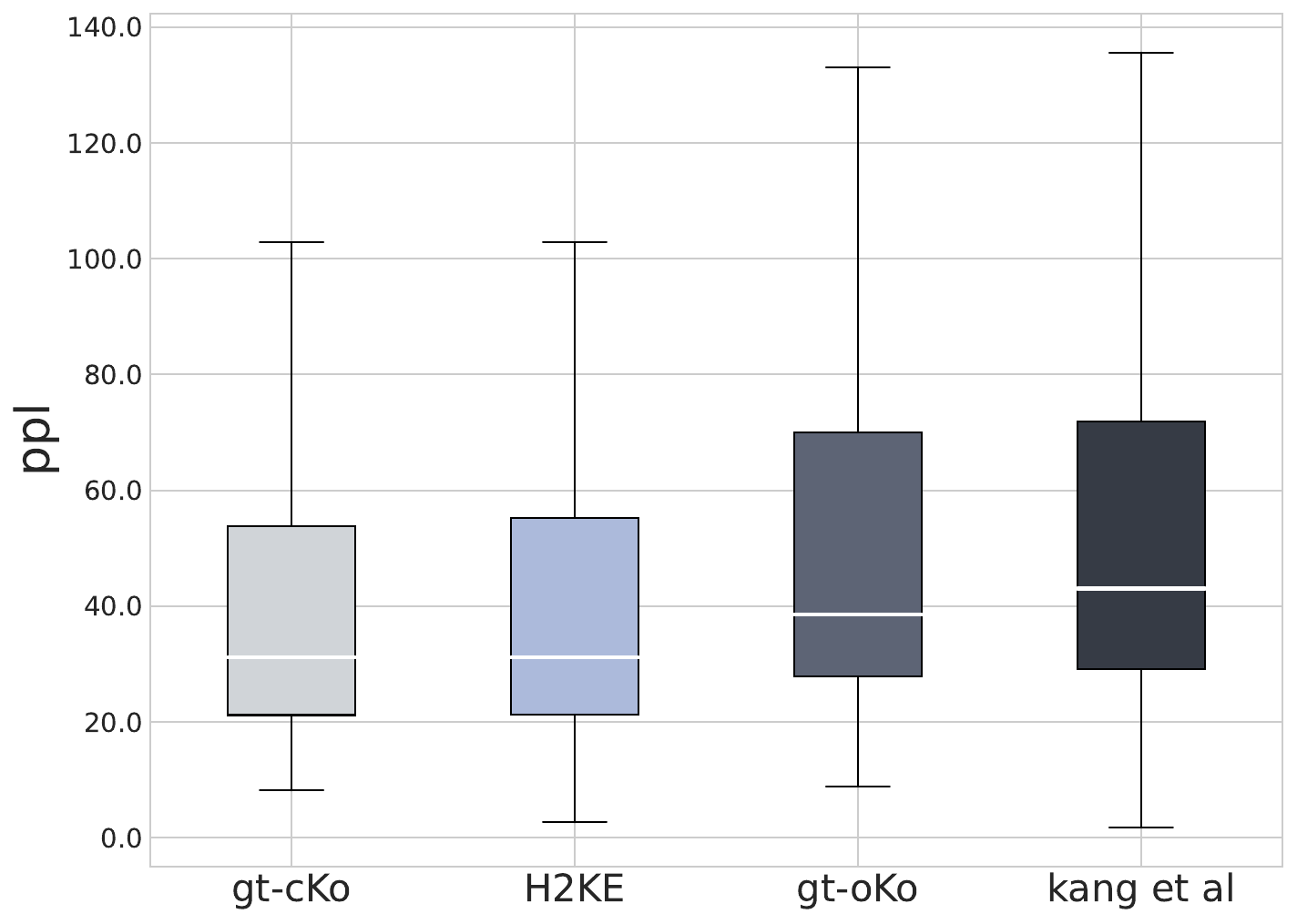}
    \caption{Per-system perplexity comparison calculated by KoGPT.}
    \label{fig:ppl_boxplot}
\end{figure}

\subsection{How contemporary is contemporary Korean translation?}

Perplexity \cite{horgan1995complexity} is the standard metric for measuring the performance of a language model, and it has been used recently to measure the deterioration of a language model over time by  \citet{lazaridou2021mind}.
To identify the difference and similarity between AJD translation, produced by different methods, and the modern Korean language, we calculate the perplexity of translations in the test set under a Korean pre-trained GPT \cite{kakaobrain2021kogpt}, and huggingface framework \cite{wolf-etal-2020-transformers}.
We used H2KE-big from Table \ref{tab:bleu} (B) in the case of the proposed approach.

\paragraph{Per-system perplexity.}

Figure \ref{fig:ppl_boxplot} draws each corpus' perplexity as a box.
There is a significant perplexity difference between the ground truth cKo (gt-cKo) and oKo (gt-oKo), which means the gt-cKo translation is closer to the modern language than the gt-oKo. 
Our generated translations result in a lower perplexity than the gt-oKo and \citet{kang2021restoring}; it is closer to the modern language similarly to gt-cKo.

\begin{table}[t]
\centering
\begin{tabular}{m{1.4cm}|llll}
\toprule
  \multicolumn{5}{c}{$P(ppl(A)<ppl(B))$} \\\midrule
 \diagbox[width=1.5cm]{(A)}{(B)}& \multicolumn{1}{p{0.8cm}}{\textbf{gt-oKo}} & \multicolumn{1}{p{0.8cm}}{\textbf{\citeauthor{kang2021restoring}}} & \multicolumn{1}{l}{\textbf{H2KE}}& \multicolumn{1}{p{0.8cm}}{\textbf{gt-cKo}} \\ \midrule
 \textbf{gt-oKo}&  &0.48*& 0.28* &0.22* \\
\textbf{\citeauthor{kang2021restoring}}  & 0.52* && 0.28* &  0.20* \\
\rowcolor[HTML]{D0D4DB}
\textbf{H2KE}& {\textbf{0.72}}*&{\textbf{0.72}}*&   &\textbf{0.54}\\
\textbf{gt-cKo}  & 0.78* & 0.80* & 0.46 & \\ \bottomrule
\end{tabular}
\caption{Pairwise perplexity comparison of each model calculated by KoGPT. Each cell shows the estimated probability of $ppl(A) < ppl(B)$ by BT model. * indicates statistically significant results with $p<0.05$.}
\label{tab:ppl_bt}
\end{table}

\paragraph{Pairwise Evaluation.} 

Because translations are associated with the same source sentences, respectively, we can compare each pair of systems by fitting Bradley-Terry (BT) model \cite{peyrard2021better, bradley1952rank}.
The BT model estimates the probability that one system is better than another based on how frequently the former system scores better.
We report the estimated probabilities, $P(ppl(A)<ppl(B))$, in Table \ref{tab:ppl_bt}.

H2KE is more like contemporary Korean than either of the ground truth oKo or \citet{kang2021restoring} with probability 0.72.
As anticipated, ground truth cKo is significantly more like contemporary Korean than both ground truth oKo and baseline.
Between H2KE and the ground truth cKo, we do not observe a significant difference in this evaluation, implying that the proposed H2KE's translations are almost on par with cKo in terms of how probable they are under a language model trained on contemporary Korean.
This observation is in agreement with our earlier observation on absolute evaluation.

\section{Human Evaluation}

We conduct human evaluation of Korean translations to confirm that H2KE's translations are both more understandable and accurate than the ground-truth oKo.
We use the Direct Assessment (DA) \cite{graham2013continuous, graham2014machine, graham2017can} as the primary method for evaluating translation systems, where
the crowd-sourced bilingual human assessors are asked to rate a translation given the source sentences by how adequately it expresses the meaning of the sentences in an analog scale \cite{akhbardeh2021findings}. 

We cannot however adopt the crowd-sourced DA approach as is because only a few historians can evaluate the meaning of translations by interpreting Hanja. 
We thus work together with ITKC and ask their experts to evaluate our generated translations according to their internal evaluation criteria. 
This is the same procedure taken to ensure the quality of human translations at ITKC.
Additionally, we conduct another evaluation to confirm whether the new Korean translation improves the understanding of historical documents for non-expert Korean speakers. 

\subsection{Expert Evaluation}

\paragraph{Evaluation Protocol.}

In ITKC, the evaluation criteria for the historical documents are divided into accuracy and fluency. Along each of these aspects, the scores are deducted according to errors that are made and the amount of deduction is determined based on the severity of each error.
In the case of accuracy, we deduct -5, -10 and -15 for word-level, phrase-level and sentence-level errors, respectively. In the case of fluency, we deduct -5 for a word-level error. 
We randomly select 45 test samples from the Annals of Jeongjo with each sample's length capped at 100 Hanja characters, for evaluation.
We ask six experts from ITKC to score both ground-truth translations as well as machine-generated translations.
Each sample is evaluated by two experts, and we report the average score. 
When there is significant disagreement between two experts, the score is adjusted through their discussion.

\begin{figure}[h]
    \centering
    \includegraphics[width=0.95\linewidth]{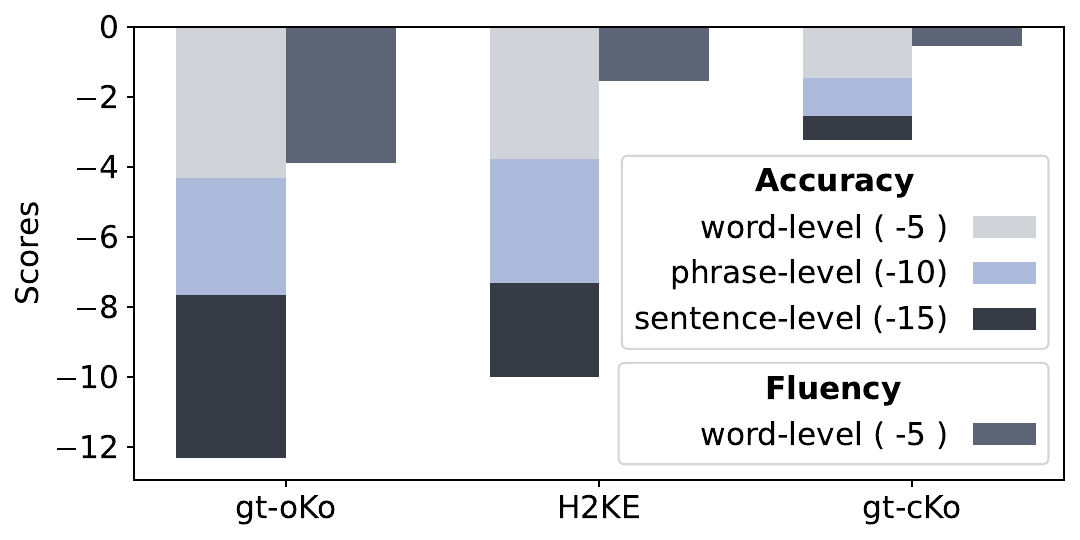}
    \caption{Average value of the deducted score per each translation by experts. Experts identified errors in the translation and subtracted scores according to the evaluation criteria.}
    \label{fig:heval_expert}
\end{figure}

\paragraph{Evaluation Result.}

Figure \ref{fig:heval_expert} shows the average deducted scores for all three cases, along both accuracy and fluency. 
As anticipated, the ground-truth cKo samples exhibit least deduction in their scores, implying that these new translations are indeed without serious translation errors and better translated.
On the other hand, the ground-truth oKo samples received most deduction in their scores, which was expected as their low readability and errors motivated re-translation of AJD in the first place. 
Our samples received worse score deduction than the ground-truth cKo, but were perceived to be significantly better than the ground-truth oKo. In particular we observed significant improvement over the original Korean translations in terms of fluency.
This outcome confirms the potential utility of the proposed approach of machine translation for re-translating the entire AJD as well as other historical Hanja documents. 

\subsection{Non-expert Evaluation}

\paragraph{Evaluation Protocol.}

To compare general public's perception of three translation types (gt-oKo, gt-cKo, and H2KE), we recruit 36 Korean speakers and request them to make pairwise comparisons of the readability.
Given a triplet \textlangle{}gt-oKo, gt-cKo, and H2KE\textrangle{} of translations of the same Hanja paragraph, we choose a random pair to give to each evaluator, either \textlangle{}gt-cKo, H2KE\textrangle{}, \textlangle{}gt-cKo, gt-oKo\textrangle{}, or \textlangle{}H2KE, gt-oKo\textrangle{}.
They have an option of `no difference,' although we encourage them to avoid it as much as possible.
We use 150 triples \textlangle{}gt-oKo, gt-cKo, H2KE\textrangle{} (450 pairs in total) from AKJ, and 150 pairs \textlangle{}gt-oKo, H2KE\textrangle{} from the annals of all the other kings (`others', in short) for which we do not have ground-truth contemporary Korean translations.
Each evaluator compares 50 pairs, and each pair is assigned three evaluators.
There are 12 different survey sheets consisting of 50 pairs each, and each survey is answered by three evaluators independently.
The details about the evaluation samples and the statistics of the evaluators are in Appendix \ref{appendix:non-expert}.

\begin{figure}[ht]
    \centering
    \includegraphics[width=0.95\linewidth]{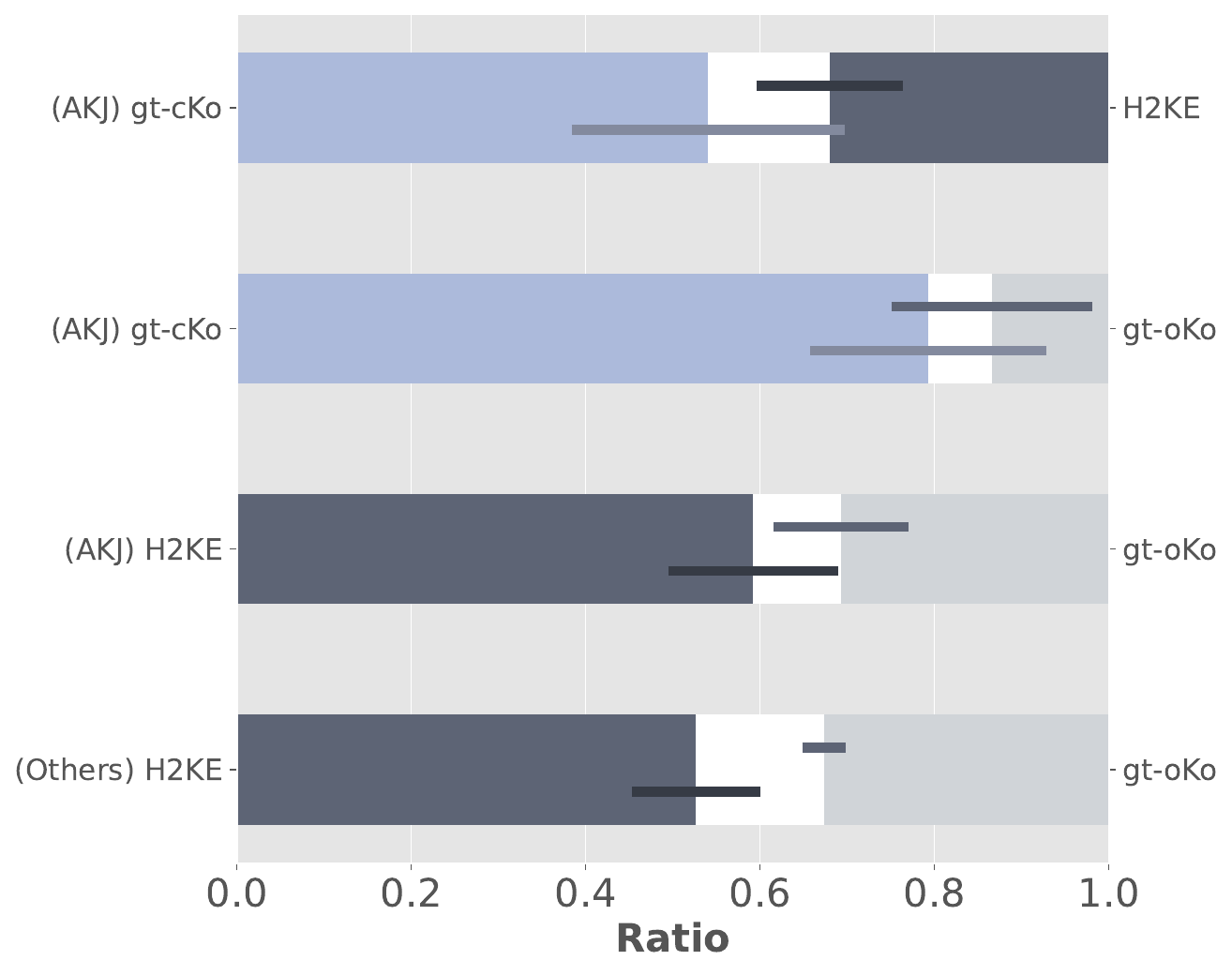}
    \caption{Result of pairwise comparison of readability by non-expert Korean speakers. The bars on each side represent the win (more understandable) rates against the other side, and the in-between white bars indicate the tie rates. Each error bar indicates the standard deviation of win rates among different survey sheets.}
    \label{fig:hevl_winrate}
\end{figure}

\paragraph{Evaluation Result.}

We use the majority vote among three evaluators' responses to decide on the winner between each pair.
When three people's opinions are divided into A, B, and no difference, we treat the pair as `no difference'.
In Figure~\ref{fig:hevl_winrate} we present the mean and the standard deviation of the win rates.

The result from AKJ shows that gt-cKo is unsurprisingly considered easier to understand than gt-oKo is, by 77.3\%.
This further emphasizes the importance and necessity of new translation of AJD for the general public.
The proposed H2KE's translations were considered more readable than oKo in AKJ by 58.0\%, which confirms the readability improvement, which was also observed with the annals of the other kings as well.
When compared against gt-cKo, gt-cKo was preferred with a probability of 52.0\%, implying that there is a room for improvement in the future.


\section{Further Analysis}
\begin{table*}[ht]
\small
\begin{tabular}{l|m{14cm}}
\toprule
\multicolumn{1}{c|}{\multirow{2}{*}{\textbf{Hanja}}}& 
\colorbox[HTML]{ACBADB}{導掌\textsuperscript{*}}之\colorbox[HTML]{D0D4D8}{科外\textsuperscript{$\dagger$}}\colorbox[HTML]{C1D1F5}{濫徵\textsuperscript{$\dagger$}}, 已極無狀, 以\colorbox[HTML]{757E94}{陳\textsuperscript{$\mathsection$}}爲起, \colorbox[HTML]{919DB8}{白地\textsuperscript{$\dagger$}}橫斂, 尤極痛駭, 使之考\colorbox[HTML]{E8E8E8}{律\textsuperscript{$\dagger$}}嚴處.
\\\cmidrule(l){2-2} 
\multicolumn{1}{l|}{}  &
Eng.) It is too bad that the \colorbox[HTML]{ACBADB}{Dojang\textsuperscript{*}} \colorbox[HTML]{C1D1F5}{excessively collected\textsuperscript{$\dagger$}} the tax \colorbox[HTML]{D0D4D8}{outside the regulations\textsuperscript{$\dagger$}}. It is even more surprising that \colorbox[HTML]{757E94}{the old land\textsuperscript{$\mathsection$}} was regarded as cultivated land and was collected for \colorbox[HTML]{919DB8}{no reason\textsuperscript{$\dagger$}}. Look at \colorbox[HTML]{E8E8E8}{the provisions of the law\textsuperscript{$\dagger$}} and let them deal with it strictly.
\\ \midrule
\multirow{1}{*}{\textbf{gt-oKo}}&\colorbox[HTML]{ACBADB}{도장(導掌)\textsuperscript{*}}이 \colorbox[HTML]{D0D4D8}{과외(科外)\textsuperscript{$\dagger$}}로 \colorbox[HTML]{C1D1F5}{남징(濫徵)\textsuperscript{$\dagger$}}하는 것은 이미 몹시 부당한 일이며 \colorbox[HTML]{757E94}{진전(陳田)\textsuperscript{$\mathsection$}}을 기경(起耕)하였다고 하여 \colorbox[HTML]{919DB8}{백지(白地)\textsuperscript{$\dagger$}}에 함부로 거두는 것은 더욱 몹시 통탄스럽고 해괴한 일이니, 그들을 \colorbox[HTML]{E8E8E8}{율(律)\textsuperscript{$\dagger$}}을 상고하여 엄히 처단하라. \\
 \midrule
\multirow{1}{*}{\textbf{gt-cKo}}  & \colorbox[HTML]{ACBADB}{도장(導掌)\textsuperscript{*}}이 \colorbox[HTML]{D0D4D8}{규정 외\textsuperscript{$\dagger$}}로 \colorbox[HTML]{C1D1F5}{지나치게 징수\textsuperscript{$\dagger$}}한 것도 대단히 형편없는 일인데, \colorbox[HTML]{757E94}{진전(陳田)\textsuperscript{$\mathsection$}}을 경작한 땅이라고 하여 \colorbox[HTML]{919DB8}{아무 근거 없이\textsuperscript{$\dagger$}} 함부로 거두었으니, 더더욱 대단히 놀랍다. \colorbox[HTML]{E8E8E8}{법률 조문\textsuperscript{$\dagger$}}을 살펴 엄히 처리하게 하라. \\
\midrule
\multirow{1}{*}{\textbf{H2KE}} & \colorbox[HTML]{ACBADB}{도장(導掌)\textsuperscript{*}}이 \colorbox[HTML]{D0D4D8}{규정 외\textsuperscript{$\dagger$}}에 \colorbox[HTML]{C1D1F5}{지나치게 징수\textsuperscript{$\dagger$}}한 것은 너무도 형편없는 짓이다. \colorbox[HTML]{757E94}{묵은 땅\textsuperscript{$\mathsection$}}을 일군 것으로 만들어 \colorbox[HTML]{919DB8}{아무런 까닭도 없이\textsuperscript{$\dagger$}} 마구 거두어들였으니, 더욱 지극히 통탄스럽고 놀랍다. \colorbox[HTML]{E8E8E8}{법률 조문\textsuperscript{$\dagger$}}을 살펴 엄히 처리하게 하라.\\
\bottomrule
\end{tabular}
\caption{The translation example of ground truth oKo, cKo, En and our generated cKo translations. 
The parenthesized words are literally translated from the original Hanja words. The same color box represents the group of words with the same semantic meaning. * indicates the proper noun; the literal translation is allowed. $\dagger$ represents the case that gt-cKo and H2KE-cKo eliminate the literal translation. $\mathsection$ is the word only our model can generate a more understandable translation.}
\label{tab:sample_unused_word}
\end{table*}

\subsection{Sample-Level Analysis of Korean Translations}

The human evaluation confirmed that H2KE significantly improves the readability and quality of the translation compared to the original oKo translations. In this section, we conduct finer-grain analysis. First,
we measure how many undesirable transliteration of Hanja words are eliminated by H2KE.
These transliterations are often marked in the corpus with their coresponding Hanja words surrouned by paranetheses.
We thus construct the archaic Hanja-based word set by extracting the gt-cKo's Hanja-based word set from the gt-oKo's. 
Among these detected transliterations, the proposed H2KE replaces 75\% with more understandable contemporary translations.

Table \ref{tab:sample_unused_word} illustrates one sample text in Hanja, ground truth oKo, cKo, and H2KE. 
The color box represents the transliterated Hanja words.
The words that have the same semantic meanings and correpond to each other across different types of translations are grouped using the same color. 
The ground truth oKo contains many literal translations, i.e. near-transliterations, identified by parentheses, and there is even a new Hanja word
(起耕)
added by the human translator. 
Compared to the gt-oKo, H2KE and gt-cKo replace most of those difficult translations with more easily understood ones. These are marked with $\dagger$. 
On the other hand, a proper noun, that is supposed to be transliterated, H2KE correctly preserves this behaviour. See Dojang
(導掌)
marked with *, which is the name of an institute.
In some cases, we notice H2KE generates a translation that is even more readable and more contemporary than the ground-truth contemporary Korean, such as the one marked as $\mathsection$.

\subsection{Sample-Level Analysis of English Translation}

Table \ref{tab:eng_ex} has an example of English translation from H2KE and Papago.
As we use the best-performing model for each case, the sample presented from H2KE and Papago are respectively translated from Hanja and oKo.
Because Papago is not aware of the historical context, it translates the word `경연' (Royal Lecture) to its homonym, a `contest.'
In contrast, our model correctly translates it into `Royal Lecture.'

\begin{table}[ht]
\centering
\begin{tabular}{@{}l|m{5.8cm}}
\toprule
\textbf{Hanja}&
隕霜.御\colorbox[HTML]{D0D4D8}{經筵}.\\
\midrule
\textbf{gt-oKo}&서리가 내렸다. \colorbox[HTML]{D0D4D8}{경연}에 나아갔다.\\
\midrule
\textbf{gt-En} & Frost appeared and the King attended the \colorbox[HTML]{D0D4D8}{Royal Lecture}.\\
\midrule
\textbf{H2KE} & Frost covered the ground. The King attended the \colorbox[HTML]{D0D4D8}{Royal Lecture}.\\
\midrule
\textbf{Papago} & It frosted. I went on to \colorbox[HTML]{D0D4D8}{the contest}.\\
\bottomrule
\end{tabular}

\caption{English translation Examples in the test set of the Annals of Sejong (4th King). Our generated sample is translated from Hanja, and the Papago sample is from ground truth oKo.}
\label{tab:eng_ex}
\end{table}

\subsection{H2KE beyond AJD}

\emph{Daily Records of the Royal Court and Important Officials} (DRRI) is another Hanja corpus, consisting of journals written in the period between the 21st King Yeongjo and the last Emperor Sunjong. 
DRRI consists of 2,329 volumes, and 42\% of the corpus has been translated manually by experts.
Unlike AJD, DRRI's original Hanja documents do not contain any punctuation marks.
This corpus is not included in the training data of our model nor that of the baseline by \citet{kang2021restoring}, which allows us to test the corpus-level generalization ability of our approach. 
We consider the translated part of DRRI after 2012 as contemporary Korean (cKo) and measure the BLEU score on this portion.

\begin{table}[ht]
\centering
\begin{tabular}{ll}
\toprule
   \textbf{Model} & \textbf{BLEU}  \\ \midrule
 \citet{kang2021restoring}& 12.96 \\
                     H2KE-oKo  & 21.50 \\ 
                     H2KE-cKo  & \textbf{32.23} \\ \bottomrule
\end{tabular}
\caption{BLEU score of translations on DRRI.}
\label{tab:add_ex}
\end{table}

We make two major observations according to the results in Table \ref{tab:add_ex}. First, H2KE-cKo produces translations that are of high quality, evident from BLEU above 30. Second, H2KE-cKo performs favourably to H2KE-oKo, which further confirms that H2KE-cKo is capable of producing translation in contemporary Korean. Finally, we observe that our approach works substantially better than the baseline, which may be due to missing punctuation marks, although we leave more detailed analysis to the future.

\section{Conclusion}
We present H2KE, a neural machine translation system for the AJD that translates from \textbf{H}anja to contemporary \textbf{K}orean and \textbf{E}nglish. 
H2KE is built on top of MNMT systems to overcome the low-resource training data problem.
H2KE shows a significantly higher BLEU score than the baseline and a current commercial translation system. 
Based on the perplexity evaluation with KoGPT, the translation samples from H2KE are closer to the contemporary Korean corpus than the ground truth original Korean translations and the baseline. 
The human evaluation results show that the translation samples from H2KE are more accurate and understandable than the ground truth original Korean. 
Finally, we translate the entire AJD to contemporary Korean and English with H2KE and publicly release the translations. 

In this work, we provide strong evidence that existing algorithms for machine translation and natural language processing generalize to a scenario where data span several centuries of an archaic language. It is highly technical in that it leads to a deeper understanding of existing algorithms and significantly extends the scope of the previous studies.


\section*{Limitations}



\emph{The Annals of Joseon Dynasty} (AJD) were written over the course of about 500 years, so naturally Hanja underwent change during long period. Capturing the temporal change would result in a better performing model.
On a related note, some entities, such as locations, and linguistic expressions may have disappeared altogether, and we simply would not be able to express those in today's language without lengthy explanations.
In the non-expert evaluation, some of the surveys reported low inter-annotator agreement because there were only three annotators per question and the evaluation of readability is subjective. The range of non-experts' prior knowledge of Korean history varies widely, and this also affects inter-annotator agreement.

\section*{Ethics Statement}

The expert evaluation was performed under Institutional Review Board (IRB) approval. 
It was conducted by the experts from the Institute for the Translation of Korean Classics (ITKC), and evaluation fees were paid to evaluators according to the ITKC's criteria for evaluation fee payment.
In recruiting non-expert evaluators, there was no discrimination against minority groups such as age, ethnicity, disability, and gender.
They were paid the compensation of more than the minimum wage of Korea.

\section*{Acknowledgements}
We would like to thank the Institute for the Translation of Korean Classics (ITKC) for providing expertise on Korean historical documents and their evaluations.
This work was partly supported by the National Research Foundation of Korea (NRF) grant funded by the Korea government (MSIT) (2022R1F1A1064401). KC was supported by Samsung Advanced Institute of Technology (under the project Next Generation Deep Learning: From Pattern Recognition to AI) and NSF Award 1922658 NRT-HDR: FUTURE Foundations, Translation, and Responsibility for Data Science.

\bibliography{emnlp2022.bib}
\bibliographystyle{acl_natbib}

\clearpage

\appendix
\section{Translation Samples}
\subsection{Annals of King Jeongjo (AKJ)}
Table \ref{tab:sample_akj} shows more examples of AKJ translated by H2KE.

\subsection{Daily Records of the Royal Court and Important Officials (DRRI)}

Table \ref{tab:sample_drri} represents the translation samples of DRRI.
The Hanja source sentences of DRRI do not contain the punctuation mark. 
The H2KE can translate the Hanja sentence to the two types of Korean, new and old Korean, by adding a different language token in front of the source sentence, so we compare both. 
The translation samples of H2KE-cKo show comparable quality to the gt-cKo, human translations. 
H2KE-oKo has the same semantic meaning as the Hanja source sentence but hurts the readability. 
The baseline model \cite{kang2021restoring} cannot generate the correct translation; a token repetition problem exists in their samples.

\section{Data Balancing Experiment}
Since our dataset consists of imbalanced types of language pairs, we experiment with the balance technology of up/down sampling proposed in \citet{liu2020multilingual}.
The result in Table \ref{tab:balancing} indicates that the up/down sampling leads to improvements in the translation to English but causes degradations in the translation to Korean.

\begin{table}[h]
\centering
\begin{tabular}{c|cc}
\toprule
 & w/o balancing & w/ balancing \\
\midrule
HJ $\rightarrow$ oKo  & 46.58 & 45.04\\
HJ $\rightarrow$ cKo  & 46.11 & 45.14\\
oKo $\rightarrow$ cKo & 45.76 & 45.25\\
HJ $\rightarrow$ EN   & 24.62 & 25.10\\
oKo $\rightarrow$ EN  & 24.59 & 25.20\\
\bottomrule
\end{tabular}
\caption{Effect of data balancing on H2KE-big. The values in `w/o balancing' column are from row (E) of Table~\ref{tab:bleu}.}
\label{tab:balancing}
\end{table}

\section{Winning Rate in Pairwise Perplexity Comparison}
Table \ref{tab:ppl_wr} represents the winning rate in pairwise perplexity comparison. 
Consistent with the BT comparison on Table\ref{tab:ppl_bt}, the translations samples from H2KE are more closer to the gt-cKo than the gt-oKo and baseline model. 
The samples that have same perplexity are exactly same, because of the short length of the source sentences.

\section{Expert Evaluation}
Table \ref{tab:eval_criteria} shows the part of the ITKC's criteria for evaluating Korean translation of historical document written in Hanja.
We directly adopt those creteria for our expert evaluation.

\begin{table}[h]
\centering
\begin{tabular}{m{1.4cm}|r|m{3.5cm}}
\toprule
\multicolumn{1}{c|}{\textbf{Error}}& \multicolumn{1}{c|}{\multirow{2}{*}{\textbf{Scale}}} & \multicolumn{1}{c}{\multirow{2}{*}{\textbf{Description}}} \\
\multicolumn{1}{c|}{\textbf{Type}} & \multicolumn{1}{c|}{}                                & \multicolumn{1}{c}{}                                      \\ \midrule
\multirow{5}{*}{Accuracy} & \multirow{2}{*}{-5}& \tabitem Mistranslation of a vocabulary\\ 
&& \tabitem Incomplete translation of a phrase\\ \cmidrule(l){2-3} 
& -10& \tabitem Mistranslation of a phrase\\ \cmidrule(l){2-3} 
                          & \multirow{2}{*}{-15}                                 & \tabitem Consecutive mistranslation of phrases                    \\
                          &                                                      &  \tabitem Mistranslation of a sentence                               \\ \midrule
\multirow{2}{*}{Fluency}  & \multirow{2}{*}{-5}                                  & \tabitem Awkward translation                                      \\
                          &                                                      & \tabitem Literal translation of unused Hanja words                 \\ \bottomrule
\end{tabular}
\caption{Evaluation criteria of ITKC for historical document translation.}
\label{tab:eval_criteria}
\end{table}

\section{Non-expert Evaluation}
\label{appendix:non-expert}

\begin{figure*}
    \centering
    \includegraphics[width=0.9\linewidth]{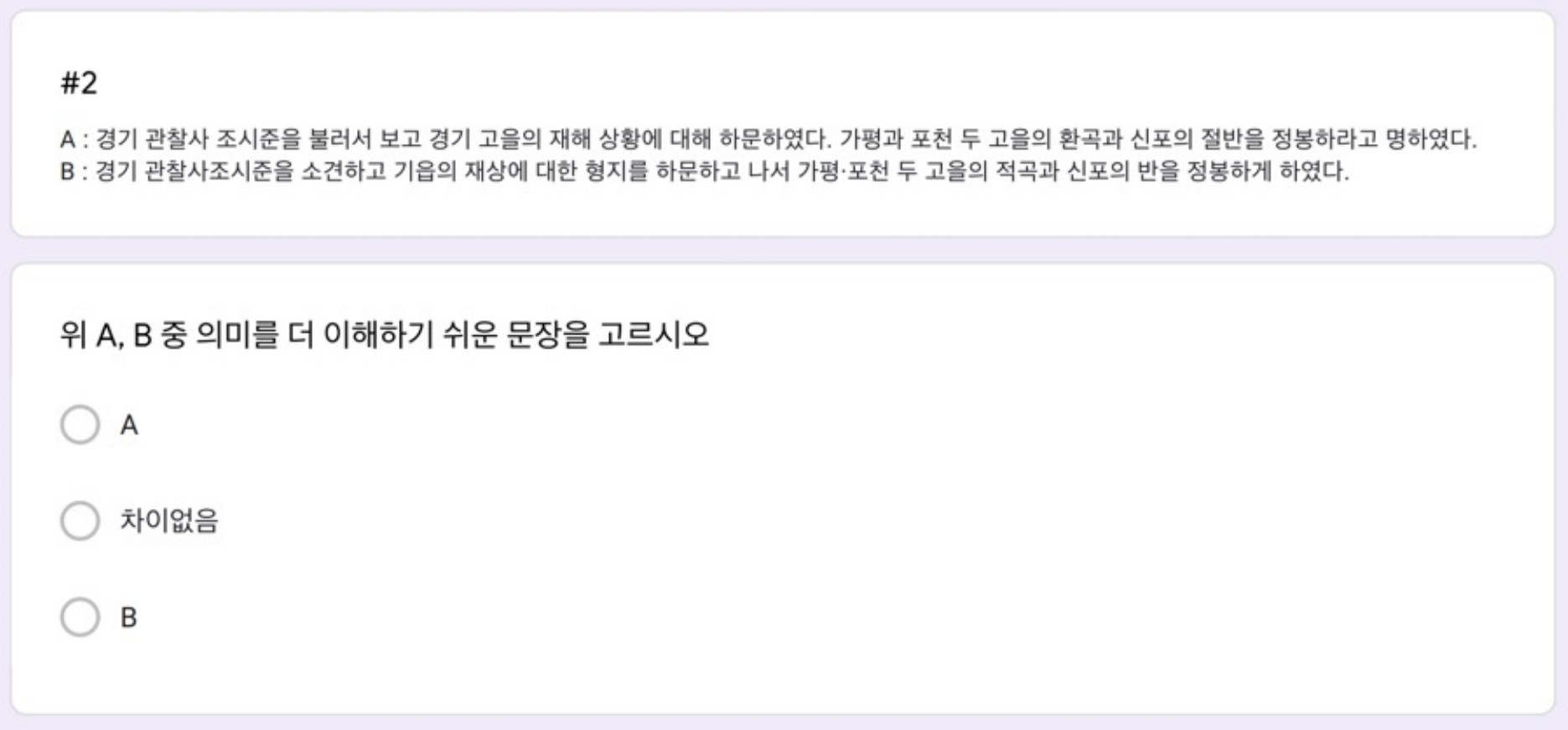}
    \caption{Screenshot of an example of non-expert evaluation. It asks to choose the more understandable one given a pair (A, B) of translations. The evaluators could choose either A, no difference, or B.}
    \label{fig:nonexp_ex}
\end{figure*}

Figure \ref{fig:nonexp_ex} shows an example question of the non-expert evaluation.
The average length of the evaluated samples is about 300 Korean letters including the spaces.
The ages of the non-expert evaluators range from 21 to 37, and the average is 24.
It implies that the evaluators are more familiar to modern Korean of the 21st century (when AJD is being newly translated) than old Korean of the 20th century (when AJD was first translated).

\begin{table*}
\renewcommand*{\arraystretch}{1.1}
\centering
\small
\begin{tabular}{cc|ccc}

\toprule
A    & B          & \multicolumn{1}{c|}{ppl(A) \textless ppl(B)(\%)} & \multicolumn{1}{c|}{ppl(A) = ppl(B) (\%)} & \multicolumn{1}{c}{ppl(A) \textgreater ppl(B)(\%)} \\ \midrule
gt-cKo  & gt-oKo        & 67.96 & 13.09 &  18.94 \\ \midrule
gt-cKo  & H2KE       & 38.71 & 25.90 &  35.37  \\
gt-cKo & \citet{kang2021restoring} &  62.39 & 13.92 & 23.67   \\   \midrule
H2KE & gt-oKo        & 68.52 & 13.92 &  17.54 \\
\citet{kang2021restoring} & gt-oKo  &  38.71 & 15.59 & 45.68   \\   \midrule
H2KE & \citet{kang2021restoring} &  61.55 & 14.20 & 24.23   \\   
\bottomrule
\end{tabular}
\caption{The winning rate in pairwise perplexity comparison of our models, ground truth samples and the baseline model.}
\label{tab:ppl_wr}
\end{table*}

\begin{table*}
\small
\begin{subtable}{\linewidth}
\begin{tabular}{@{}l|m{14cm}}
\toprule
\multicolumn{1}{@{}l|}{\multirow{2}{*}{\textbf{Hanja}}}& 卜相. 拜判敦寧徐命善爲右議政, 金尙喆·鄭在謙陞爲領左相. \\
\cmidrule(l){2-2} 
\multicolumn{1}{l|}{}  &
Eng.) [The king] nominated candidates for the State Council. He appointed Seo Myeong-seon, the Magistrate of Donnyeongbu, to the Right State Councilor, and promoted Kim Sang-cheol and Jeong Jon-gyeom to the Chief State Councilor and the Left State Councilor. \\
\midrule
\multirow{1}{*}{\textbf{gt-oKo}}& 복상하였다. 판돈녕서명선을 우의정에 제배하고 김상철·정존겸을 올려서 영상과 좌상으로 삼았다. \\
 \midrule
\multirow{1}{*}{\textbf{gt-cKo}}  & 의정 후보를 뽑았다. 판돈녕부사 서명선을 우의정에 제수하고, 김상철과 정존겸의 좌차를 영의정과 좌의정으로 올렸다. \\
\midrule
\multirow{1}{*}{\textbf{H2KE}} & 의정의 후보를 뽑았다. 판돈녕부사 서명선을 우의정에 제수하고, 김상철과 정재겸을 승진시켜 영의정과 좌의정으로 삼았다. \\
\bottomrule
\end{tabular}
\caption{}
\end{subtable}

\begin{subtable}{\linewidth}
\begin{tabular}{@{}l|m{14cm}}
\toprule
\multicolumn{1}{@{}l|}{\multirow{2}{*}{\textbf{Hanja}}}& 兩司啓請: “逆禶孥籍, 依金吾草記擧行, 啓能亟施孥籍之典.” 不允. \\
\cmidrule(l){2-2} 
\multicolumn{1}{l|}{}  &
Eng.) Both Offices (the Office of the Censor General and the Office of the Inspector General) said, “we ask to apply the law to make wife and children as slaves and confiscate family property on the traitor Lee Chan as in the document from the State Tribunal, and enforce the law as soon as possible on Hong Gye-neung as well,” but it was not granted.\\
\midrule
\multirow{1}{*}{\textbf{gt-oKo}}& 양사에서 아뢰기를, “역적 이찬의 노적을 금오의 초기대로 거행하고, 홍계능에 있어서도 시급히 노적하는 법을 시행하기를 청합니다.” 하였으나, 윤허하지 아니하였다. \\
 \midrule
\multirow{1}{*}{\textbf{gt-cKo}}  & 양사가 아뢰어, 역적 이찬에 대해 처자식을 노비로 삼고 가산을 몰수하는 법을 의금부의 초기대로 거행할 것과 홍계능에 대해서도 속히 처자식을 노비로 삼고 가산을 몰수하는 법을 시행하도록 청하니, 윤허하지 않았다.
 \\
\midrule
\multirow{1}{*}{\textbf{H2KE}} & 양사가 아뢰어, 역적 이찬에 대해 처자식을 노비로 삼고 가산을 몰수하는 것을 의금부의 초기대로 거행하고 홍계능에 대해 처자식을 노비로 삼고 가산을 몰수하는 법을 속히 시행할 것을 청하였는데, 윤허하지 않았다. \\
\bottomrule
\end{tabular}
\caption{}
\end{subtable}

\caption{Translation samples of the Annals of King Jeongjo (AKJ).}
\label{tab:sample_akj}
\end{table*}

\begin{table*}
\small
\centering
\begin{subtable}{\linewidth}
\centering
\begin{tabular}{@{}l|m{12cm}}
\toprule
\multicolumn{1}{@{}l|}{\multirow{2}{*}{\textbf{Hanja}}}& 政院以李萬秀方在罷散啓稟敎以敍用 \\
\cmidrule(l){2-2} 
\multicolumn{1}{l|}{}  &
Eng.) When the Royal Secretariat asked the king that Lee Mansu was currently in bankruptcy, the king asked to hire him.\\
\midrule
\multirow{1}{*}{\textbf{gt-cKo}}& 정원이 이만수가 현재 파산 상태에 있다고 계품하여, 전교하기를,"서용하라."하였다.\\
 \midrule
\multirow{1}{*}{\textbf{H2KE-cKo}}& 정원이, 이낙수가 현재 파산 중에 있다고 주상에게 여쭈니, 서용하라고 하교하였다. \\
\midrule
\multirow{1}{*}{\textbf{H2KE-oKo}} & 정원에서 이유수가 바야흐로 파산에 있다는 것으로 계품하니, 서용하라고 하교하였다. \\ \midrule
\multirow{1}{*}{\citet{kang2021restoring} } &  정원에서 이만수가 현재 파산 계품교를 아뢰니, 서용하여 서용하였다. \\
\bottomrule
\end{tabular}
\caption{}
\end{subtable}

\begin{subtable}{\linewidth}
\centering
\begin{tabular}{@{}l|m{12cm}}
\toprule
\multicolumn{1}{@{}l|}{\multirow{2}{*}{\textbf{Hanja}}}& 入來時用吉服行禮故雖爲吹打而旣是二十七朔之內則禮罷之後不可作軍樂卽令停止 \\

\cmidrule(l){2-2} 
\multicolumn{1}{l|}{}  &
Eng.) When they came in, they played Chwita (musical instrument) put on the Gilbok (casual clothes worn after a funeral), but they should not play military music after it is over because it is within a 27-month period. Let it stop immediately.\\
\midrule

\multirow{1}{*}{\textbf{gt-cKo}}& 들어올 때 길복을 입고 예를 행하기 때문에 취타를 하였지만 27개월 의 거상 기간 안이니 예가 끝난 후에 군악을 연주해서는 안 된다. 즉시 정지하게 하라.\\
 \midrule
\multirow{1}{*}{\textbf{H2KE-cKo}}& 들어올 때에는 길복을 입고 예를 행하기 때문에 취타하더라도 27개월 이내에야 하니, 예를 마친 뒤에는 군악을 연주하지 말고 즉시 정지하게 해야 합니다. \\
\midrule
\multirow{1}{*}{\textbf{H2KE-oKo}} &  들어올 때에 길복을 입고 예를 행하기 때문에 취타하기는 하였으나, 이미 27개월 안이므로 예를 파한 뒤에 군악을 만들어서 곧 멈추게 할 수 없다 합니다. \\ \midrule
\multirow{1}{*}{\citet{kang2021restoring} } &  칙사가 돈화문 뒤에 규례대로 취타하면 부칙사가 말하기를, 「칙사가 정지한다.’고 말하기를, ‘칙사가 돈화문 뒤에 규례대로 취타를 한다. \\
\bottomrule
\end{tabular}
\caption{}
\end{subtable}

\begin{subtable}{\linewidth}
\centering
\begin{tabular}{@{}l|m{12cm}}
\toprule

\multicolumn{1}{@{}l|}{\multirow{2}{*}{\textbf{Hanja}}}& 今年兩西畿內之民當疲於使星支應通計勅行使行儐使往來道臣行部則爲二十一次之多三道之民其何以堪乎如有 \\

\cmidrule(l){2-2} 
\multicolumn{1}{l|}{}  &
Eng.) This year, people in Yangseo and Gyeonggi Province will be tired of entertaining envoys. 
If you calculate the total number of visits of the envoy and the procession of officials throughout the jurisdiction, there are 21 times, so how can the people of the three provinces handle it?\\ \midrule

\multirow{1}{*}{\textbf{gt-cKo}}& 금년에 양서와 경기의 백성은 사신을 지응하느라 지쳤을 것이다. 칙사의 행차, 사신의 행차, 빈사가 왕래하는 것과 도신이 관내를 순행하는 것을 통틀어 계산해 보면 21차례나 되니 세 도의 백성이 어떻게 감당할 수 있겠는가.\\
 \midrule
\multirow{1}{*}{\textbf{H2KE-cKo}}& 올해 양서와 경기 지역의 백성들은 사신을 접대하는 데 지쳐 있을 것이다. 칙사 일행의 빈사가 왕래하는 도신의 행부를 통틀어 계산하면 21차례나 되니, 3개 도의 백성들이 어떻게 견디겠는가. \\
\midrule
\multirow{1}{*}{\textbf{H2KE-oKo}} &  올해 양서와 기내의 백성들은 의당 사성을 지응하고 칙사 일행의 빈사를 접대하러 왕래하는 도신의 행부를 통계하는 데 지쳐야 할 것인데, 23일이나 되는 3도의 백성들이 어떻게 견디겠는가.  \\ \midrule
\multirow{1}{*}{\citet{kang2021restoring} } &
올해 금년 양서의 기내의 백 백성이 사성의 지응과 지응과 사성 지응과 지응해야 할 때에 관사 행사 행사 와 사행 사행의 사행 사행사 가 왕래가 왕래사와 사행에 왕래하는 도신 의 행부 는 21차의 많은 3도의 백성들이 어떻게 감당할 수 있겠는가. 만일 한 분분의 폐폐를 제거하는 도리가 있으면 삼도의 백성이 한 분의 분의 분수를 위하여 백성을 위하여 백성을 위해야 할 수 있겠다고 생각할 수 있겠는가.
\\
\bottomrule
\end{tabular}
\caption{}
\end{subtable}

\caption{The translation samples of DRRI.}
\label{tab:sample_drri}
\end{table*}


\end{document}